\documentclass[11pt]{article}

\usepackage[preprint]{acl}

\usepackage{times}
\usepackage{latexsym}

\usepackage[T1]{fontenc}
\usepackage[utf8]{inputenc}

\usepackage{microtype}

\usepackage{inconsolata}

\usepackage{graphicx}

\usepackage{booktabs, multirow, enumitem}
\usepackage[most]{tcolorbox}
\usepackage[capitalize]{cleveref} % Automatically capitalizes "Figure"

\setlist{
  itemsep=1pt,     % Space between consecutive items
  topsep=1pt,      % Space before the list starts and after it ends
  parsep=0pt,      % Space between paragraphs within a single item
  partopsep=0pt    % Extra space added if the list is preceded by a blank line
}

\crefformat{section}{\S#2#1#3}
\Crefformat{section}{\S#2#1#3}

\crefname{figure}{Figure}{Figures}
\Crefname{figure}{Figure}{Figures}

\title{Do Large Language Models Always Tell The Same Stories?}

\author{Thennal D K \and Hans Ole Hatzel \\
  Language Technology Group \\
  University of Hamburg \\
  \texttt{thennal10@gmail.com}, \texttt{hans.ole.hatzel@uni-hamburg.de}}

\begin{document}
\maketitle
\begin{abstract}
Recent advances in large language models (LLMs) have enabled the generation of high-quality prose, yet whether these models are capable of generating diverse or creative artifacts remains a contested question. In this work, we investigate the diversity of LLM-generated stories through the framework of \textit{narrative similarity}. Using a contrastive framework and a dataset of human-written stories and prompts from \texttt{r/WritingPrompts}, we collect narrative similarity judgments across 10 representative LLMs, utilizing both human evaluations and three different automatic annotation methods. Our findings reveal a clear trend: LLM-generated narratives are consistently more similar to each other than human-written stories are. We demonstrate that frontier models in particular converge on a ``mean'' generic narrative that approximates individual human stories but lacks the collective diversity of human authors. Finally, we show that common mitigation strategies, including negative prompting and temperature scaling, fail to meaningfully address this homogeneity.
\end{abstract}

\section{Introduction}

With recent advances, large language models (LLMs) are being increasingly applied to creative writing tasks, from short stories to poems, with the ability to craft personalized and interactable narratives \citep{telekiSurveyLLMsStory2025,yuanWordcraftStoryWriting2022,tianAreLargeLanguage2024}. Frontier LLMs are now capable of generating stories and prose that evaluators often find more pleasing compared to human-written alternatives \citep{klinge2026llms, chakrabartyCanGoodWriting2026}. Despite these capabilities, whether LLM-generated artifacts are as \textit{creative} or \textit{diverse} in comparison to humans remains an open question, with often contradictory results in the literature \citep{gilhoolyAIVsHumans2024,bellemare-pepinDivergentCreativityHumans2026,luAIHumanitysSalieri2025,padmakumarMeasuringLLMNovelty2025,marcoSmallLanguageModels2025,shypula2025evaluating}.

\begin{figure}[t]
    \centering
    \includegraphics[width=1\linewidth]{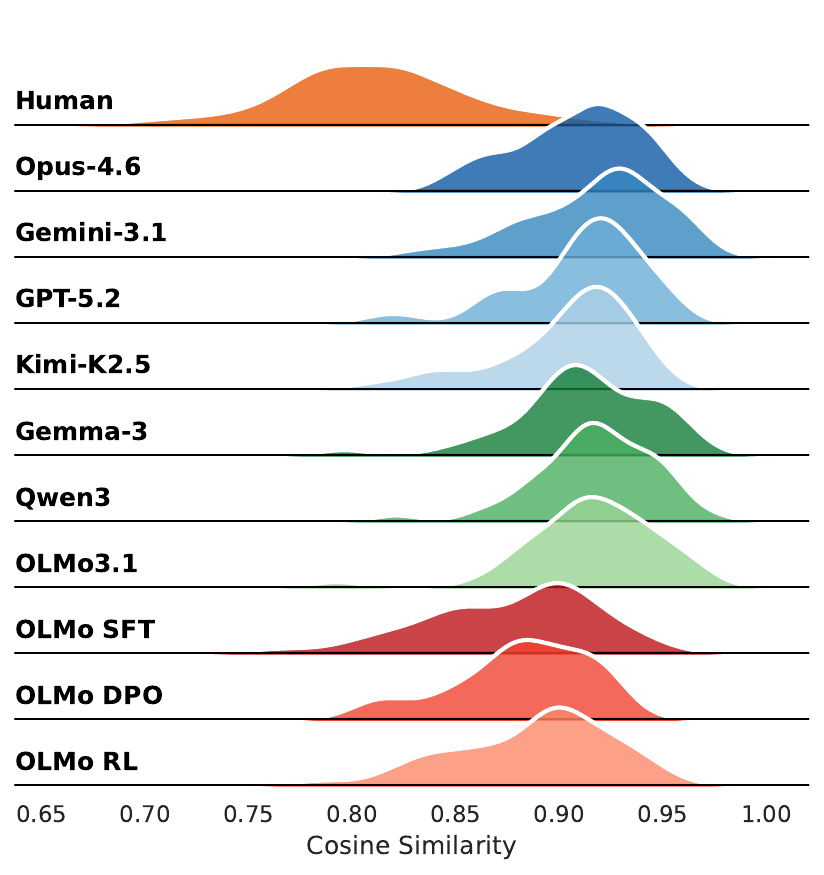}
    \caption{The distribution of narrative component embedding similarity between pairs of stories, generated by the same model or written by humans, across our dataset. We note that human-written stories are consistently less similar to each other than LLM-generated ones.}    
    \label{fig:sim_distribution}
\end{figure}

We suspect that the breadth of contradictory findings is directly tied to the ambiguity in operationalizing creativity, particularly as it applies to machine generation. Recent work by \citet{luRethinkingCreativityEvaluation2026} evaluates the methods used in the field and finds that directly judging the creativity of LLM-generated text is prone to error and inconsistency. Within this background, our work narrows the focus to the distinct but related question of \textbf{narrative diversity}: How similar are LLM-generated narratives, within and between models as well as in comparison to human-written ones? 

To address this question, we utilize a contrastive annotation framework based on narrative similarity \citep{hatzelSemEval2026Task42026}. By asking annotators to determine which of two candidate stories are narratively closer to a reference story, we build a set of paired similarity judgments that allow us to make inferences on the narrative diversity of stories within a given pool. To build the pools, we collect human-written prompts and stories from \texttt{r/WritingPrompts}\footnote{\url{https://www.reddit.com/r/WritingPrompts/}}, and use the prompts to generate corresponding LLM narratives, allowing us to make a direct comparison. 

Through human evaluation and three different automated annotation methods, we conduct a large-scale exploration of narrative diversity across 10 LLMs, encompassing closed-source frontier models, smaller open-source models, and a sequence of post-training checkpoints. As illustrated in \cref{fig:sim_distribution}, our findings reveal a consistent and stark difference in the narrative diversity of humans and LLMs. Across model families, scale, and post-training, LLM-generated narratives are overwhelmingly judged to be more similar to other LLM-generated narratives, with particularly high rates of similarity within stories generated by the same model. Closed-source LLMs are able to emulate human-written narratives but lack the diversity between them, while smaller LLMs generate narratives that are notably distinct from humans yet still homogeneous. Finally, we test common mitigation techniques such as negative prompting and temperature scaling, finding that they do little to improve diversity.\footnote{Our code and data are available at: \url{https://github.com/thennal10/narrative-similarity}.}

\section{Related Work}
As generative language models explode in popularity and usage, studies have tried to quantify their creativity in various domains, from problem solving to creative drawing \citep{nathPencilsPixelsSystematic2025, tianMacGyverAreLarge2024, caoEvaluatingTextCreativity2026}. We broadly categorize approaches into two groups: rubric-based evaluation, and automated lexical metrics.

Several works adapt well-known psychometric tests and human-evaluated rubrics for creativity, such as the Alternative Uses Task \citep{gilhoolyAIVsHumans2024}, Torrance Test of Creative Thinking \citep{chakrabartyArtArtificeLarge2024}, and the Divergent Associations Test \citep{bellemare-pepinDivergentCreativityHumans2026}. Some find that LLMs outperform the average human baselines in these psychometric measures \citep{gilhoolyAIVsHumans2024, bellemare-pepinDivergentCreativityHumans2026}, while others note that they are far from the creativity expressed in expert human writing \citep{chakrabartyArtArtificeLarge2024, gomez-rodriguezConfederacyModelsComprehensive2023}. While specific psychometric tests have been shown reliable for humans, whether they are equally applicable for LLMs remains contested \citep{nakajimaDivergentCreativityHumanBased2026}. To scale these evaluations, some recent work heavily relies on LLM-as-a-Judge setups \citep{paech2023eq,marcoSmallLanguageModels2025}, but given the subjective nature and difficulty of the task, these approaches are prone to inconsistency and bias \citep{luRethinkingCreativityEvaluation2026}.

Other works opt for lexical metrics to quantify creativity. \citet{luAIHumanitysSalieri2025} utilizes a metric derived from n-gram novelty, finding that professional authors score $66.2\%$ higher than LLMs, while alignment reduces the score considerably. However, \citet{padmakumarMeasuringLLMNovelty2025} utilizes a similar metric based on unseen and task-specific n-grams, finding that post-training methods actually improve novelty, beyond the average human baseline. \citet{shypula2025evaluating} also utilizes a framework measuring a derived semantic diversity, and comes to a similar conclusion regarding the positive effects of post-training and scale. 
%\citet{jiangArtificialHivemindOpenEnded2025} also utilize text-based overlap to find high inter-model and intra-model similarity, but notably do not compare with human writing. 
In general, \citet{saakyanDeathNoveltyNGram2025} and \citet{luRethinkingCreativityEvaluation2026} caution against n-gram novelty as a proxy for creativity, finding that it does not match expert evaluations.

%In general, as \citet{luRethinkingCreativityEvaluation2026} outlines in their exploration of existing creativity measures, these automated approaches suffer from key limitations: perplexity-based metrics reflect fluency rather than novelty, n-gram metrics primarily measure lexical diversity, and LLM-as-a-Judge produces inconsistent judgments with minor prompt variations.
 
\citet{xuEchoesAIQuantifying2025} and \citet{tianAreLargeLanguage2024} come closest to directly tackling our question of narrative diversity. \citet{tianAreLargeLanguage2024} compares LLM-generated plot summaries with Wikipedia film summaries, and categorizes them into 7 story arcs, finding that LLM-generated plots overrepresent certain arcs. \citet{xuEchoesAIQuantifying2025} utilize stories from \texttt{r/WritingPrompts} and Wikipedia plot summaries to define a \textit{Sui Generis} score that quantifies the repetition in plot elements, finding that LLMs may ``echo'' particular plot elements. However, both approaches consider narrative within a narrow scope; \citet{tianAreLargeLanguage2024} only considers a fixed categorization of one aspect of the narrative, and \citet{xuEchoesAIQuantifying2025} focuses on plot element repetition within 30-word segments in alternate continuations of stories. Neither evaluates the narrative as a whole, and both utilize condensed plot summaries rather than the narratives contained within an organic medium. Our work diverges by directly comparing the overarching narratives that underpin model-generated stories, and provides broader results on the effects of post-training and mitigation techniques.
% Their score seems to be overly complex and essentially penalize any segment that doesn't introduce a new plot element, which I would argue isn't actually indicative of a lack of diversity, but could simply be a result of cohesiveness.

\section{Methodology}
%Prior research has showcased that directly judging the creativity or diversity of LLM-generated narratives is prone to error, both in terms of operationalizing creativity as well as in the measurement itself \citep{luRethinkingCreativityEvaluation2026}. Thus, we consider a related but distinct research question: How similar are LLM-generated narratives, within and between models as well as in comparison to human-written ones?

We pose the following central research question: How similar are LLM-generated narratives, within and across models as well as in comparison to human-written ones? In order to tackle this directly, we borrow the contrastive format used in a recent shared task on \textbf{narrative similarity} \citep{hatzelSemEval2026Task42026}. Given a pool of stories from a set of generators, we take a triplet: a reference story, and candidate stories A and B. We format this triplet as an annotation task, asking the annotator to choose whether the narrative of story A or story B is closer to the reference. 

\begin{tcolorbox}[colback=cyan!5!white,colframe=cyan!75!black]
  We use the term \textit{generator} to indicate the entity that created a given story; Humans or the specific LLM.
\end{tcolorbox}

We focus on short narratives written with an explicit writing prompt, a choice that aligns our approach with prior work and makes large scale annotations feasible. The inclusion of a writing prompt provides a starting point for the LLM generation without explicitly dictating the narrative structure, and closer reflects the practical usage of LLMs in story writing \citep{yuanWordcraftStoryWriting2022}. 

Our setup provides several distinct advantages: a contrastive format is easier for annotators to judge, and less bound by the specific guidelines for judgment \citep{hatzelSemEval2026Task42026}. It is less subjective than prior attempts asking annotators to score creativity based on a rubric, while still measuring the narrative diversity of a generator via narrative similarity judgements. Finally, restricting ourselves to stories written with a specific writing prompt reduces the difficulty of the judging task further, and provides a fine-grained comparison. 

\subsection{Visualization}
\label{sec:visualization}

Our setup measures narrative diversity in a contrastive framework: similarity judgments directly depend on the pool of generators and there is conceptual difference between comparisons where a generator is the reference and where it is a candidate. Any inference thus requires a broader overview of the data that we could not encompass with a single per-generator metric. We instead turn to visualizations, specifically in the form of the heatmap.

In the heatmap, each row $i$ corresponds to the reference generator, and each column $j$ corresponds to the selected candidate generator (i.e. the generator whose story is judged to be more similar). Each element $M_{i,j}$ represents the normalized selection rate of generator $j$. Specifically, given a triplet with the reference story generated by $i$ and one of the two candidate stories generated by $j$, $M_{i,j}$ is the proportion of such triplets where the story generated by $j$ was found to be more similar to the reference. In effect, large values in the matrix indicate that the corresponding generators’ stories are more often narratively similar. %For clarity, if the judgments were made at random, we would expect all elements to be $~50\%$, and if generator $j$ was always judged to be more similar, we would expect $M_{i,j}$ to be $100\%$ (and correspondingly, if a generator was always judged to be less similar, the value would be $0\%$). 

\section{General Setup}
\label{sec:experiments}

In this section, we provide the general outline of our setup, including the models, triplet generation strategy, and dataset we use. 
\subsection{Dataset}
\label{sec:dataset}

In order to provide a direct comparison between human and LLM-generated narratives, we require a dataset of story prompts and corresponding stories written by humans. Following related work, we source stories from Reddit's \texttt{r/WritingPrompts} subreddit, a forum dedicated to amateur story writing. Users post writing prompts and other users comment with stories based on the prompt, providing a direct analogy to prompting an LLM. Users can interact via upvotes and downvotes resulting in a score that indicates a given story's popularity.

To quantify the similarity between human-written stories, we require some triplets where the reference and one of the two candidate stories were written by humans, and so we need two human stories per prompt. To ensure baseline quality and prevent length bias, we filter for stories between 200 and 300 words with at least 100 upvotes. To mitigate exposure bias, where authors might alter their submission after reading earlier ones, we exclusively select story pairs posted within 15 minutes of each other. Following manual cleanup of paratext and duplicates, our dataset comprises 44 prompts, each associated with two pairs of human stories. This provides a diverse set of story prompts that enables a large set of combinatorial cross-generator comparisons.

\subsection{Models}
\label{sec:model_pools}
To evaluate the effects of model architecture, scale, and post-training, we select 10 total models from three different categories:

\begin{itemize}
    \item \textbf{Closed-source models:} A selection of state-of-the-art models to evaluate frontier LLMs, with Claude Opus 4.6 \citep{anthropic2026claude46}, Gemini 3.1 Pro \citep{deepmind2026gemini31}, GPT-5.2 \citep{singhOpenAIGPT5System2026}, and Kimi K2.5 \citep{teamKimiK25Visual2026}.
    \item \textbf{Open-source models:} Open-source models in the $\sim$30 billion parameter range, with Gemma 3 27B \citep{teamGemma3Technical2025a}, OLMo 3.1 32B Instruct \citep{olmoOlmo32026}, and Qwen3 30B \citep{yangQwen3TechnicalReport2025}.
    \item \textbf{OLMo models:} The OLMo 3 7B Instruct checkpoints \citep{olmoOlmo32026} at various stages of post-training, specifically after supervised fine-tuning model (SFT), direct preference optimization (DPO), and reinforcement learning (RL). We omit the base model as it did not generate coherent stories and consistently failed to stay within the word limit in early experiments.
\end{itemize}

A list of the specific model identifiers are provided in \cref{sec:exact_models}. For each writing prompt in our dataset, we prompt each model twice, creating two generations for each prompt-model combination. %Each story pool is thus constructed of two stories per model per prompt, in addition to the two human stories per prompt.

\subsection{Triplet Generation}
\label{sec:triplet_selection}

Within a given pool of stories, we generate triplets such that all stories are used as the reference exactly once, and all combinations of generators are included in the candidates for each reference. This scheme generates 2112 triplets with 4 generators, and 4400 triplets with 5 generators, and 53,240 triplets with 11 generators. For details, see \cref{sec:appendix:triplet_selection}.

\section{Annotation}
\label{sec:annotation}

To evaluate narrative similarity across our generated stories, we rely on annotated triplets. We first conduct human evaluation on a subset of the data, providing preliminary insights. Subsequently, we explore and validate three automated annotation methods against this baseline, allowing us to scale our evaluation. We will release all our annotations as well as the code and model checkpoints for our automated annotation methods.

\subsection{Human Annotations}
\label{sec:humaneval}

\begin{table}[ht]
\centering
\begin{tabular}{lccc}
\toprule
\textbf{Category} & \textbf{N} & \textbf{Pref.} & \textbf{Agr.} \\
\midrule
\texttt{(H, H, M\textsubscript{1})} & $11$ & \texttt{M\textsubscript{1}} ($72.7\%$) & $100.0\%$  \\
\texttt{(H, M\textsubscript{1}, M\textsubscript{2})} & $11$ & $-$ & $72.7\%$  \\
\texttt{(M\textsubscript{1}, H, M\textsubscript{1})} & $7$ & \texttt{M\textsubscript{1}} ($100\%$) & $100.0\%$  \\
\texttt{(M\textsubscript{1}, H, M\textsubscript{2})} & $28$ & \texttt{M\textsubscript{2}} ($91.1\%$) & $89.3\%$  \\
\texttt{(M\textsubscript{1}, M\textsubscript{1}, M\textsubscript{2})} & $18$ & \texttt{M\textsubscript{1}} ($88.9\%$) & $88.9\%$ \\
\texttt{(M\textsubscript{1}, M\textsubscript{2}, M\textsubscript{3})} & $25$ & $-$ & $76.0\%$  \\
\midrule
Overall & $100$ & $-$ & $86.0\%$  \\
\bottomrule
\end{tabular}
\caption{Inter-annotator agreement (Agr.) for each triplet category. The total number of triplets (N), and the preferred candidate (Pref. ) as well as the percentage of annotations it was preferred in is also provided, except for categories where candidates cannot be distinguished with respect to the reference.}
\label{tab:human_eval_agreement}
\end{table}

To establish a ground-truth baseline, we first annotate triplets via human evaluation on 100 sampled triplets. In addition to humans, we pick two models from each of the model categories described in \S\ref{sec:model_pools} in order to include all types of LLMs used in our study while keeping the total generator pool reasonably small. Specifically, we opt for Opus-4.6, Kimi-K2.5, Gemma-3, Qwen3, OLMo-SFT, and OLMo-DPO, for a total of 6 models. In order to ensure sufficient comparisons with human-written stories, we opt for a stratified sampling procedure that oversamples them. Annotations were provided by annotators in India and Nigeria, with two independent annotations per triplet. We utilize the expert guidelines from \citet{hatzel_2025_16537509} with minimal modification. Annotators were also asked to report their confidence in their selection on a scale of 1 to 5. Additional details are provided in \cref{sec:human_eval_details}.

As there are not enough annotations for a full heatmap, we abstract specific generators within the triplets and aggregate them into triplet categories of the form \texttt{(X, Y, Z)}, where \texttt{X}, \texttt{Y}, and \texttt{Z} are generators. We use \texttt{H} to represent human generated stories and \texttt{M\textsubscript{k}} for the k-th unique generator model in that triplet. We do not distinguish by the ordering of the candidates, e.g. \texttt{(\texttt{M\textsubscript{1}}, \texttt{H}, \texttt{M\textsubscript{2}})} = \texttt{(\texttt{M\textsubscript{1}}, \texttt{M\textsubscript{2}}, H)}. For details on this notation scheme, see \cref{sec:appendix:triplet_categories}.

Table \ref{tab:human_eval_agreement} reports the inter-rater agreement between the two annotator cohorts, as well as the preferred candidate in each category. Overall agreement is high at $86\%$, and we note that agreement remains high for categories with repeated generators or a human-generated candidate. %, i.e. \texttt{(H, H, M\textsubscript{1})}, \texttt{(M\textsubscript{1}, H, M\textsubscript{1})}, \texttt{(M\textsubscript{1}, H, M\textsubscript{2})}, and \texttt{(M\textsubscript{1}, M\textsubscript{1}, M\textsubscript{2})}. 
Conversely, the agreement is particularly low when the candidate stories are generated by two different models, distinct from the generator, i.e. \texttt{(H, M\textsubscript{1}, M\textsubscript{2})} and \texttt{(M\textsubscript{1}, M\textsubscript{2}, M\textsubscript{3})}. This indicates that these categories are more ambiguous and subjective, and annotation is not as reliable. 
%However, we note that these are also categories which are least relevant for our inferences. Specifically, which of two LLMs is more similar to a human reference is largely inconsequential, beyond noting that certain LLMs may emulate human narrative structures more than other models. In the same vein, determining which of two distinct LLMs is narratively closer to a third LLM provides us little information on self-similarity or in contrast with human stories. 
We also note that this also aligns with our hypothesis: LLMs generate similar stories, and thus are harder to distinguish between as candidates to a reference that is completely distinct (human) or equally similar (another LLM). 

The preferred candidate for each category further complements our hypothesis. When evaluated against model-generated reference, a model candidate is preferred over a human candidate $91.1\%$ of the time if the models differ, and $100\%$ of the time if the model is the same. Even with a human reference and a human candidate, a model candidate is preferred $72.7\%$ of the time, indicating that human stories are less similar to each other than it is to a model story. %We opt to deemphasize the \texttt{(H, M\textsubscript{1}, M\textsubscript{2})} category as they are less relevant.

Regardless, with only 100 triplets, broader inferences and comparisons between specific generators are difficult to make, and thus we turn to automated annotation.
\subsection{Automated Annotation}
\label{sec:judge}

Due to the number of annotations required for a comprehensive overview, we require automated annotation. Inspired by the approaches outlined in \citet{hatzelSemEval2026Task42026}, we investigate three distinct methods:
\begin{itemize}
    \item \textbf{LLM-as-a-Judge}: Utilizing GPT-5 to directly annotate triplets based on a given categorization of narrative similarity \citep{paech2023eq, hatzelSemEval2026Task42026}. We also prompt for a confidence score in the judgment on a scale of 1 to 5, as with human annotation.
    \item \textbf{Narrative Component Embedding}: Decomposing stories into narrative components (abstract themes, actions, and outcomes) via GPT-5, embedding them using Gemini Embedding, and using the cosine similarity to annotate triplets \citep{hatzelSemEval2026Task42026,leeGeminiEmbeddingGeneralizable2025}. 
    \item \textbf{Preference Model}: A lightweight preference model based on the Bradley-Terry formulation \citep{bradleyRankAnalysisIncomplete1952} trained on synthetically generated annotations, with Qwen3 1.7B as the base model \citep{yangQwen3TechnicalReport2025}. 
\end{itemize}

The specific prompts used and other additional details for each setup are given in \cref{sec:appendix_judge}.

\begin{table}[t]
\centering
\begin{tabular}{lccc}
\toprule
\textbf{Category} & \textbf{LLM} & \textbf{Emb.} & \textbf{PM} \\
\midrule
\texttt{(H, H, M\textsubscript{1})} & $81.8\%$ & $81.8\%$ & $72.7\%$ \\
\texttt{(H, M\textsubscript{1}, M\textsubscript{2})} & $68.2\%$ & $50.0\%$ & $50.0\%$ \\
\texttt{(M\textsubscript{1}, H, M\textsubscript{1})} & $100.0\%$ & $100.0\%$ & $100.0\%$ \\
\texttt{(M\textsubscript{1}, H, M\textsubscript{2})} & $91.1\%$ & $87.5\%$ & $91.1\%$ \\
\texttt{(M\textsubscript{1}, M\textsubscript{1}, M\textsubscript{2})} & $83.3\%$ & $88.9\%$ & $83.3\%$ \\
\texttt{(M\textsubscript{1}, M\textsubscript{2}, M\textsubscript{3})} & $68.0\%$ & $80.0\%$ & $64.0\%$ \\
\midrule
Overall & $81.0\%$ & $82.0\%$ & $77.0\%$ \\
\bottomrule
\end{tabular}
\caption{Average agreement with human annotators of the LLM-as-a-Judge (LLM), narrative component embedding (Emb.), and preference model (PM) annotation methods across the different triplet categories.}
\label{tab:automated_annotation_agreement}
\end{table}

\begin{figure*}[ht]
    \centering
    \includegraphics[width=1.0\linewidth]{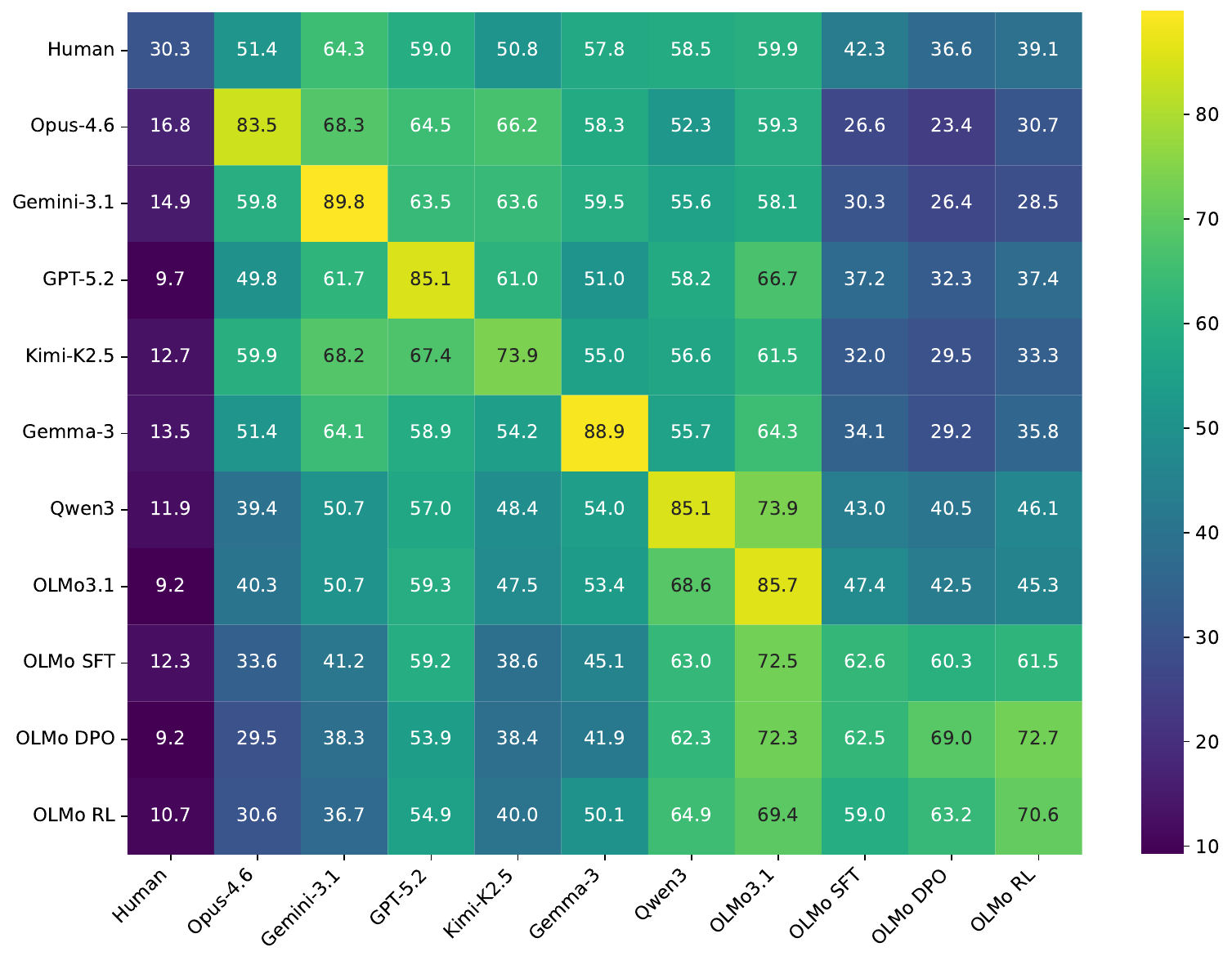}
    \caption{Similarity heatmap showcasing the normalized selection rate for all models annotated via narrative component embedding. The rows indicate the reference and the columns indicate the selected candidate generator, as detailed in \cref{sec:visualization}.}
    \label{fig:heatmap_main_all}
\end{figure*}

\begin{figure*}[ht]
    \centering
    \includegraphics[width=1.0\linewidth]{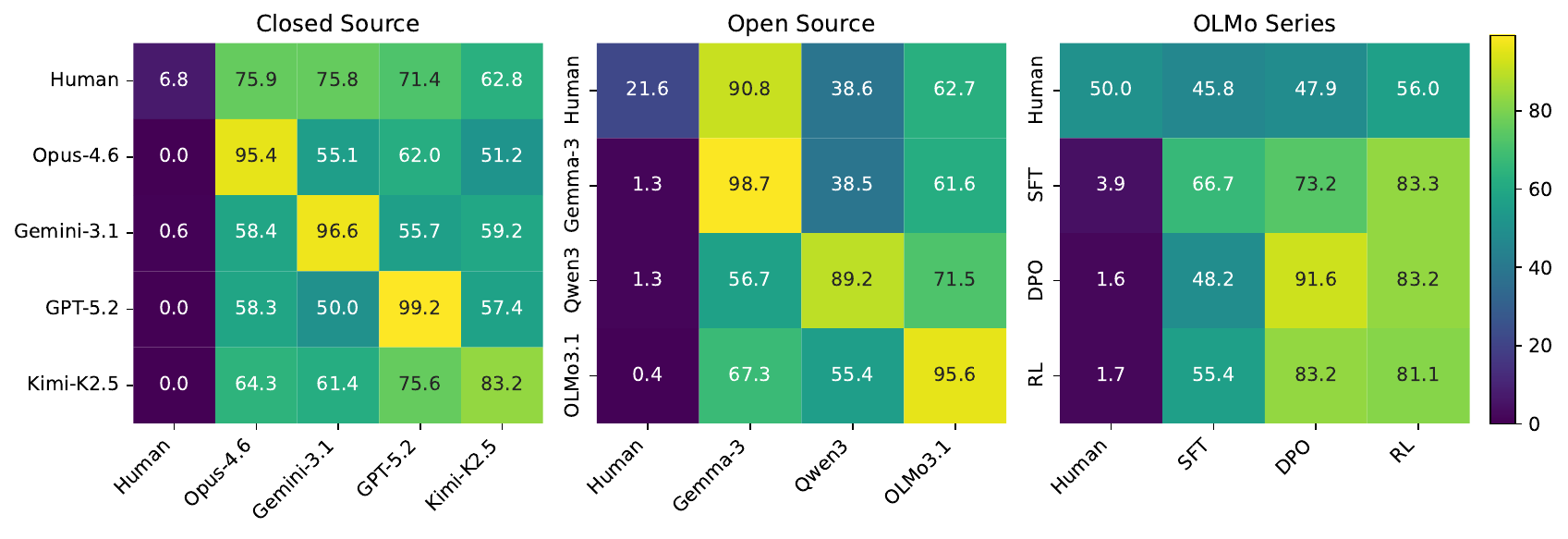}
    \caption{Similarity heatmap of triplets annotated by the LLM judge and with a confidence filter of 5 within the closed-source, open-source, and OLMo generator pools.}
    \label{fig:heatmap_main_highconf}
\end{figure*}

We evaluate the agreement of all three automated annotation methods against the human annotations, with the results compiled in \cref{tab:automated_annotation_agreement}. LLM-as-a-Judge and narrative component embedding perform equally well, with the latter performing slightly better overall at $82\%$ agreement. The preference model performs notably worse at $77\%$ agreement. We note that the two categories in which judges performed the worst are the same categories where inter-annotator agreement is notably lower. The preference model's lower overall agreement is also largely a result of poorer performance in these categories. %As we detailed in \cref{sec:humaneval}, these categories are the two which are least relevant for our inferences, and so the lower agreement is less concerning. 

Furthermore, we find that the LLM judge's self-reported confidence serves as a reliable indicator of accuracy. Overall agreement reaches $92.8\%$ when the LLM reports maximum confidence (a score of 5), suggesting a confidence cutoff can effectively filter reliable annotations (a full heatmap and confidence breakdown is provided in \cref{sec:appendix:human_llm_confidence}).

Given that all three methods largely performed well, we opt for each in different contexts: for large-scale comparisons with the entire model pool, we use the narrative component embedding method, for a more detailed comparison between specific generators, we use the LLM-as-a-Judge method, and for additional mitigation experiments where efficient repeated annotations are required, we use the preference model. 

\section{Narrative Similarity}
\label{sec:results_main}

Using all three automated annotation methods, we conduct a large-scale evaluation of narrative similarity across model pools.

\subsection{All Models}
\label{sec:results_main:all_models}

First, we use narrative component embedding to annotate triplets generated for all 11 generators (10 models and humans). The resulting heatmap is provided in \cref{fig:heatmap_main_all}. 

We note a universal trend: LLM-generated stories were judged to be substantially more similar to other LLM-generated stories, with only a small fraction (between $9.2\%$ and $16.8\%$) of LLM-generated stories judged to be more similar to a human-written one. Additionally, self-similarity (the diagonal elements) is high across all models, and in particular the open and closed source models: almost all in the $80\%$ to $90\%$ range with the exception of Kimi-K2.5 at $73.9\%$. In contrast, human self-similarity largely remains low at $30.8\%$.%, with human-generated reference stories judged to be more similar to an LLM-generated one than another human-authored story. %We thus make a clear inference from these results:

\begin{tcolorbox}[colback=green!5!white,colframe=teal!75!black,title=Key Takeaway]
  LLM-generated stories are narratively more similar to each other than human-written stories are.
\end{tcolorbox}

The OLMo 7B models follow the same general trends, but diverge in notable ways. In particular, self-similarity is not substantially higher than inter-model similarity, while similarity with other models is noticeably low: a non-OLMo 7B model-generated story is judged to be more similar to an OLMo 7B model-generated one $26.6\%$ to $47.4\%$ of the time. Given that the OLMo 7B models are all checkpoints from the same base model, this is not a surprising result. Nevertheless, we note that out of all models, the OLMo SFT model has the lowest self-similarity at $62.6\%$.

\subsection{Within Model Categories}
\label{sec:results_main:model_categories}

We take a more fine-grained look at each model category using the LLM judge. The closed source, open source, and OLMo models are considered separately in their own generator pools. Within each pool, we select and annotate triplets, filtering out all annotated triplets with a confidence score less than 5, as lower-confidence annotations are significantly less reliable. The resulting heatmaps are provided in \cref{fig:heatmap_main_highconf}. 

As before in \cref{sec:results_main:all_models}, we observe that LLM-generated stories are judged to be substantially more similar to other LLM-generated stories than to human-generated ones, but with starker difference: across all three pools, model-human similarity is between $0.0\%$ and $3.9\%$. Human self-similarity is also extremely low in the closed source pool at $6.8\%$, but increases to $21.6\%$ in the open source pool, and to $50.0\%$ in the OLMo pool. 

We may interpret this as a trend towards converging on a ``mean'' narrative with increasing scale and post-training. Large closed-source models generate stories which approximate human-generated narratives better but lack the diversity found between different human narratives. Therefore, a human-generated reference is likelier to be similar to a model-generated story than they are to another human-generated story. In contrast, the OLMo series generates narratives which are distinctly different from human-generated narratives, and so a human-generated reference is equally dissimilar to both model and human-generated candidates. We again note that the OLMo SFT model has the lowest self-similarity among all models at $66.7\%$, while the DPO and RL models have both high self-similarity and inter-modal similarity, forming a distinct visual block. %This suggests that post-training also directly contributes to the lack of narrative diverstiy. 
%Ultimately, these results point to another key inference:

\begin{tcolorbox}[colback=green!5!white,colframe=teal!75!black,title=Key Takeaway]
With increasing scale and post-training, LLMs converge on generic narratives, closer to individual human narratives while lacking the diversity found across them.
% Frontier LLMs converge on generating generic narratives which are closer to any one human-generated narrative, but lack the diversity found between different human narratives.
\end{tcolorbox}

We note that similar results were observed without the confidence filter as well, as shown in \cref{sec:appendix:model_categories}. We also annotate the same model pools with the preference model in \cref{sec:appendix:model_categories_bt}, showing that there is little difference between the results of the two annotation methods.

\begin{figure*}[ht]
    \centering
    \includegraphics[width=1.0\linewidth]{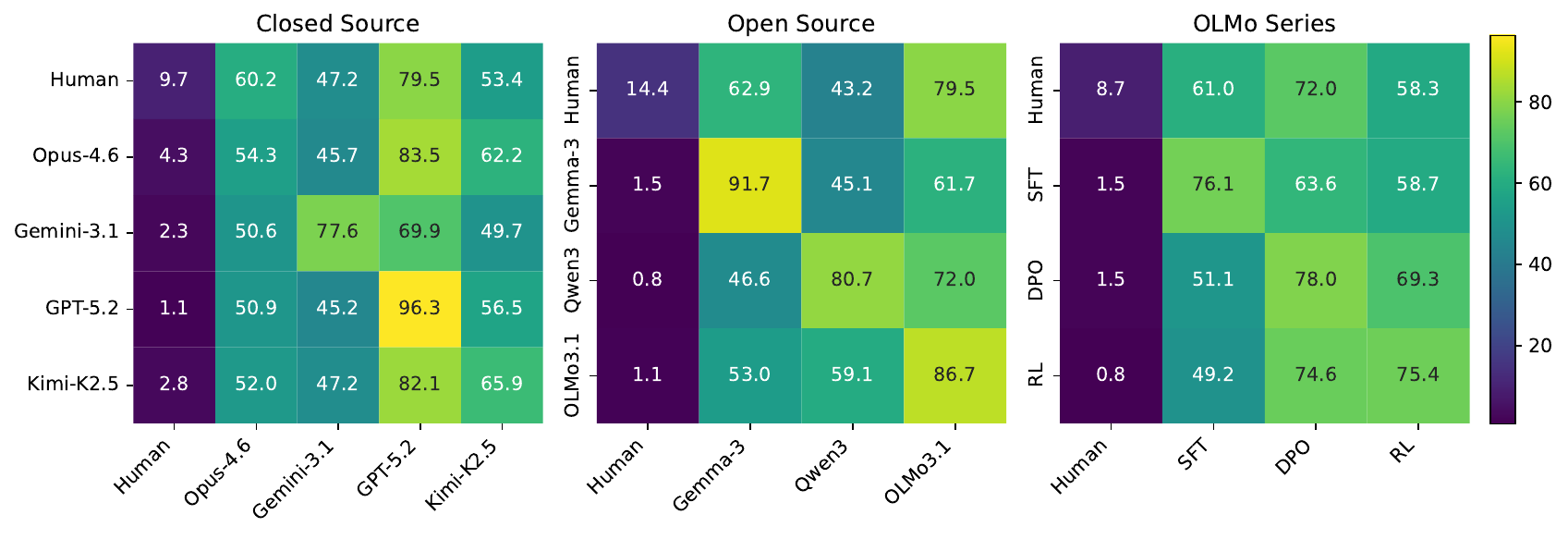}
    \caption{Similarity heatmap of triplets annotated by the preference model within the closed-source, open-source, and OLMo generator pools, and with stories generated sequentially (each model is provided all prior generated stories).}
    \label{fig:heatmap_seq_bt}
\end{figure*}

\begin{figure}[ht]
    \centering
    \includegraphics[width=1.0\linewidth]{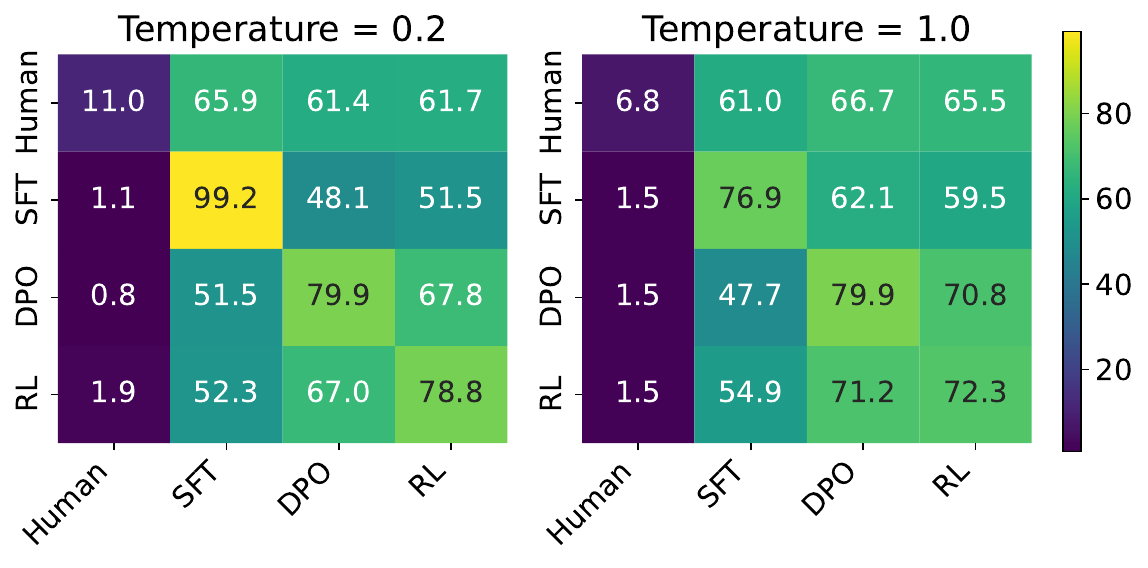}
    \caption{Similarity heatmap of triplets annotated by the preference model, within the OLMo pool under a selected range of temperatures.}
    \label{fig:heatmap_temp_sweep}
\end{figure}

\section{Mitigation Strategies}

Several mitigation strategies exist in the literature to increase diversity, including prompting and decoding methods \citep{zhangVerbalizedSamplingHow2025b,tianAreLargeLanguage2024,parkAvoidanceDecodingDiverse2025a}. We test whether these have any effect on narrative diversity using the preference model.

\subsection{Sequential Negative Prompting}

A common and simple mitigation strategy is to generate stories sequentially, providing the previous story as context and prompting the model to generate a different story \citep{zhangVerbalizedSamplingHow2025b}. We test a more rigorous version of this strategy for all three pools: instead of only providing the previous stories as context for that specific model, we generate stories within the pool in a round-robin fashion, appending \textbf{all} prior model-generated stories (including other models within the pool) and explicitly prompting the model to generate a story that is as different as possible. We annotate the resulting triplets with the preference model, and the resulting heatmap is provided in \cref{fig:heatmap_seq_bt}. We note that while self-similarity is reduced, it is still largely higher than inter-model similarity, and LLM-generated stories remain substantially more similar to each other than to human-generated stories, with little improvement in overall diversity. We also note that GPT-5.2-generated stories are particularly likely to be judged more similar regardless of the reference, indicating that this strategy's effectiveness is also dependent on the specific model used. %Ultimately, even when LLMs are aware of other stories within the pool and prompted to generate dissimilar ones, this still does little to boost overall diversity. 

\begin{tcolorbox}[colback=green!5!white,colframe=teal!75!black,title=Key Takeaway]
    Even with explicit negative prompting against prior generated stories, LLM-generated narratives remain homogeneous. 
\end{tcolorbox}

\subsection{Sampling Parameters}
\label{sec:results_mitigation:sampling_params}

We test the effects of two common sampling parameters that are utilized to make model outputs more diverse: temperature and top-P \citep{holtzmanCuriousCaseNeural2019,ackleyLearningAlgorithmBoltzmann1985}. Using the preference model and OLMo pool, we generate stories under all combinations of temperature $T\in \{0.2, 0.4, 0.6, 0.8, 1.0\}$ and top-P $P\in \{0.85, 0.9, 0.95, 1.0\}$. Surprisingly, we find that top-P has no consistent impact. For brevity, we showcase the heatmaps of three selected temperature values ($0.2, 0.6, 1.0$) all with top-P of $0.9$ in \cref{fig:heatmap_temp_sweep}. The full set of heatmaps can be found in \cref{sec:appendix:param_sweep}.

We note that the only model that seems to be appreciably affected by sampling parameters is SFT, with self-similarity decreasing considerably from $99.2$ to $76.9$ as temperature goes from $0.2$ to $1.0$. However, SFT-generated stories are still rarely judged to be more similar to a human-generated story, even with a temperature of $1.0$. These results also support prior work, which have found that post-training leads to a collapse in the logit probability calibration, making temperature scaling ineffective \citep{tanRestoringExplorationPostTraining2026,xieCalibratingLanguageModels2024}.

\begin{tcolorbox}[colback=green!5!white,colframe=teal!75!black,title=Key Takeaway]
    Sampling parameter scaling can reduce self-similarity in non-aligned models but have little effect in preference-tuned models. 
\end{tcolorbox}

\section{Conclusion}

Our study provides a novel and robust operationalization of creativity through the lens of narrative diversity. By employing a contrastive similarity framework within the domain of writing prompt stories, we provide a fine-grained comparison of narratives within and across humans and LLM models. Utilizing both human and automated annotation methods, our findings reveal that across a wide range of models, LLMs generate narratives that are consistently more similar to each other than human-written stories. We also find that post-training and increasing scale both contribute to this lack of diversity, with frontier LLMs in particular generating generic ``mean'' narratives that are closer to human-generated stories but lacking the diversity found between different human-generated stories. 

We also test common mitigation strategies to improve diversity, and find that none of them show meaningful improvements, pointing to a clear question for future work. Within this conception of creativity, we conclude that today's LLMs are not creative storytellers, and this limitation cannot be ignored as they see increasing use in creative applications. 

\section*{Limitations}

As we source our writing prompts and human-written stories from Reddit in the timespan of 2014 to 2024, the training data of the models we evaluate may be contaminated. However, we posit that contamination is unlikely to be a relevant factor in our findings, as we would expect high similarity between human-written stories and model-generated stories if the models had memorized or otherwise utilized elements of the human-written stories.

Additionally, when training the preference model, we exclude the specific stories in human-annotated triplets from the training data. However, the stories are still generated by the same models that we evaluate with, and with largely the same set of writing prompts. Thus, the preference model may not be generalizable beyond these writing prompts or model pools. 

Finally, while we test the effects of sampling parameters on diversity, we do not test decoding methods such as avoidance decoding \citep{parkAvoidanceDecodingDiverse2025a}, which may have a stronger effect. Regardless, as we are primarily interested in narrative-level diversity, we hypothesize that token-level decoding methods are unlikely to be an effective mitigation strategy.

\section*{Ethical Considerations}

As our work is largely concerned with analysis of existing LLMs, we do not foresee any direct ethical concerns with our study. The artifacts we release, including the annotations, code, and model checkpoints, are all explicitly designed for narrative similarity evaluation; they do not contain sensitive, toxic, or personally identifiable information, nor are they suited for generating harmful content. Thus, we do not anticipate any potential for misuse.

\bibliography{custom}

\appendix

\section{Models}
\label{sec:exact_models}

We utilize the following list of models throughout our experiments, with the corresponding OpenRouter API endpoint (for the proprietary models) or the Hugging Face repository name (for all other models) provided:

\paragraph{Closed-Source Pool}
\begin{itemize}
    \item \textbf{Opus-4.6}: \texttt{anthropic/claude-opus-4.6}
    \item \textbf{Gemini-3.1}: \texttt{google/gemini-3.1-pro-preview}
    \item \textbf{GPT-5.2}: \texttt{openai/gpt-5.2}
    \item \textbf{Kimi-K2.5}: \texttt{moonshotai/kimi-k2.5}
\end{itemize}

\paragraph{Open-Source Pool}
\begin{itemize}
    \item \textbf{Gemma-3}: \texttt{google/gemma-3-27b-it}
    \item \textbf{OLMo3.1}: \\ \texttt{allenai/Olmo-3.1-32B-Instruct}
    \item \textbf{OLMo3.1-Think}: \texttt{allenai/Olmo-3.1-32B-Think}
    \item \textbf{Qwen3}: \texttt{Qwen/Qwen3-30B-A3B}
\end{itemize}

\paragraph{OLMo 7B Series}
\begin{itemize}
    \item \textbf{Base}: \texttt{allenai/Olmo-3-1025-7B}
    \item \textbf{SFT}: \texttt{allenai/Olmo-3-7B-Instruct-SFT}
    \item \textbf{DPO}: \texttt{allenai/Olmo-3-7B-Instruct-DPO}
    \item \textbf{RL}: \texttt{allenai/Olmo-3-7B-Instruct}
\end{itemize}

We use GPT-5 (\texttt{gpt-5}) as the judge for all LLM-as-a-Judge evaluations. Unless specified otherwise, all stories are generated with the default sampling parameters for each model, or with a temperature and top-P of $1.0$ when the default is not specified. All prompts used to generate stories are provided in \cref{prompt:sequential_generation}.

\section{Licenses}

\subsection{Dataset}

The stories and prompts from \texttt{r/WritingPrompts} are publicly available on Reddit, and the original authors retain the copyright to their respective texts. In compliance with Reddit's Terms of Service regarding data redistribution, we do not release the raw text of the stories or prompts, but provide a complete list of post and comment IDs used in our dataset. All annotations are released under the CC BY-SA 4.0 License, and all code and model checkpoints are released under the MIT License.

\subsection{Models}
Our evaluation utilizes a combination of open-source and proprietary LLMs. The proprietary models are subject to their respective commercial API Terms of Service. The open-source models are subject to the following licenses, which provide permission for research and non-commercial use:
\begin{itemize}
    \item \textbf{OLMo 3 and 3.1 Series:} Apache 2.0 License.
    \item \textbf{Gemma 3:} Gemma Terms of Use.\footnote{https://ai.google.dev/gemma/terms}
    \item \textbf{Qwen3:} Apache 2.0 License.
\end{itemize}

\section{Triplet Generation}
\label{sec:appendix:triplet_selection}

For a pool of stories $S$, we construct triplets $(s_r, s_a, s_b)$ by iterating through every story in the pool to serve as a reference story, $s_r \in S$. For a given $s_r$ associated with prompt $p$, we iterate through all combinations of generators $\{a,b\} \in {G \choose 2}$. As there are two stories for each generator and prompt, we randomly select one for $a$ and $b$, providing us $s_a$ and $s_b$. To ensure the triplet consists of distinct stories, we enforce $s_a \neq s_r$ and $s_b \neq s_r$. We randomize $s_a$ and $s_b$ to limit the effects of positional bias, and the resulting generated triplets for specific pools are kept consistent across experiments. In total, $|S| \times {|G| \choose 2}$ triplets are selected in a particular pool.

\section{Human Evaluation}
\label{sec:human_eval_details}

In this section, we provide additional details on the human evaluation setup.

\subsection{Sampling Triplets}
\label{sec:human_eval_sampling}

We utilize a stratified sampling procedure to generate the human evaluation triplets. As we are specifically interested in comparisons that include human-written stories, our sampling strategy explicitly upsamples those triplets. To construct a triplet, we first randomly select a writing prompt and perform a weighted sampling for a reference generator $r$, assigning a weight of $3$ to humans and $1$ to all others. We then sample candidate generators $a$ and $b$, ensuring that at least $30\%$ of the time either $a$ or $b$ is human; for the remaining $70\%$ of cases, $a$ and $b$ are sampled uniformly at random without replacement. Finally, one of the two stories per prompt per generator is uniformly sampled, yielding $(s_r,s_a,s_b)$. We note that one of the writing prompts (and all stories corresponding to it) was excluded from this process, as it was used for an example in the guidelines given to the annotators.

\subsection{Annotation Setup}

Annotations were collected via an online annotation platform (omitted for anonymity). To account for potential cultural or linguistic biases in narrative interpretation, we sourced annotators from two distinct geographic cohorts: India and Nigeria. Both regions possess a large population of fluent English speakers but offer diverse cultural lenses through which narrative tropes and similarities might be perceived. 

As mentioned in \cref{sec:dataset}, we remove all metadata and paratext, including any identifying information about the authors, and only retain the text of the prompts and stories. The \texttt{r/WritingPrompts} subreddit rules\footnote{https://www.reddit.com/r/WritingPrompts/wiki/rules} explicitly forbid harmful content, and we additionally conduct an automated check with Gemini 3.1 Pro across both the human and model-generated stories.

Every triplet was annotated once by both cohorts, resulting in two independent human judgments per triplet. The expert guidelines used are provided in full in \cref{prompt:narrative_similarity_guidelines_1} and \cref{prompt:narrative_similarity_guidelines_2}. Annotators were compensated at an hourly rate of \$19.5 (\$1.3 per triplet with an estimated annotation time of 4 minutes per triplet), well above the median wage in both regions \citep{ilo2024globalwage}.

\subsection{Triplet Categories}
\label{sec:appendix:triplet_categories}

In aggregate, the exact models utilized often matters less than the nature of the comparison---for instance, we care more about whether a human reference is being compared to two model-generated candidates, or whether a model reference is being compared to another generation from the same model and a generation from a different model, than we do about which specific models are being compared. 
Thus, when reporting certain statistics such as annotator agreement, we abstract the specific generator identities within each triplet into broader categories. 

Specifically, for a given triplet, each corresponding generator is converted to either \texttt{H}, denoting human, or \texttt{M\textsubscript{k}} (where \texttt{k} $\in \{1, 2, 3\}$), denoting a model. The subscript \texttt{k} serves to distinguish distinct models within the triplet. The order of the candidate stories do not matter; for example, \texttt{(M\textsubscript{1}, H, M\textsubscript{2})} and \texttt{(M\textsubscript{1}, M\textsubscript{2}, H)} are both categorized the same, and denoted as the former.

To illustrate, the category \texttt{(H, M\textsubscript{1}, M\textsubscript{2})} isolates all instances where a human reference is compared against two different models. Similarly, the category \texttt{(M\textsubscript{1}, M\textsubscript{1}, M\textsubscript{2})} isolates self-similarity comparisons, where a model reference is compared to another generation from the same model, and a generation from a different model. By partitioning the data in this manner, we can independently assess statistics across structurally distinct categories of the task.

\subsection{Human and LLM Judge Confidences}
\label{sec:appendix:human_llm_confidence}

\begin{figure}[ht]
   \centering
   \includegraphics[width=\linewidth]{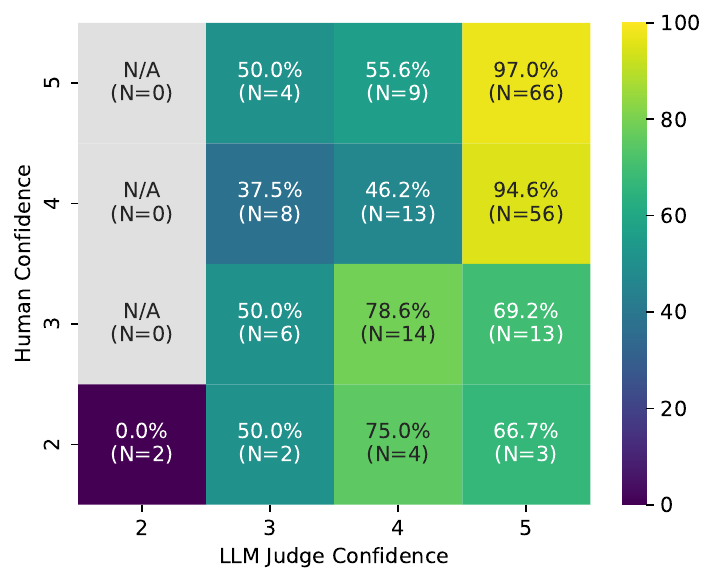}
   \caption{Average LLM judge agreement with human annotators split by reported confidence. Neither the LLM judge nor the human annotators ever reported a confidence score of 1.}
   \label{fig:confidence_heatmap}
\end{figure}

As both human annotators and the LLM judge provide confidence scores, we compile a heatmap showcasing average agreement across confidence scores in \cref{fig:confidence_heatmap}. We note that while the confidence scale ranged from 1 to 5, neither the LLM judge nor the human annotators ever provided a confidence score of 1, corresponding to the case where both stories are equally similar to the reference. For the cases where the LLM judge chose a confidence score of 2-3, the agreement is essentially at random regardless of annotator confidence. Interestingly, when the LLM judge selects a confidence score of 4, agreement is moderate when human annotator confidence is low (2-3), but reverts to random chance when human annotator confidence is high (4-5). Nonetheless, for the majority of samples, the LLM judge selects a confidence score of 5. In these cases, agreement is particularly high when human annotators are also confident, with $94.6\%$ agreement with human annotator confidence $4$ and $97.0\%$ with human annotator confidence $5$. In the rarer cases where human annotators have low confidence, agreement is low but above random chance, at $66.7\%$ for a human annotator confidence of $1$. Overall, the LLM judge agrees with annotators $92.8\%$ when it selects a confidence score of 5, which it does for $69$ of the triplets. This suggests that a confidence cutoff of $5$ can be used to improve the reliability of the LLM judge annotations, though it is likely that will bias the resulting annotation set towards the easier triplets.

\section{Automated Annotation}
\label{sec:appendix_judge}

\begin{figure*}[ht]
    \centering
    \includegraphics[width=1.0\linewidth]{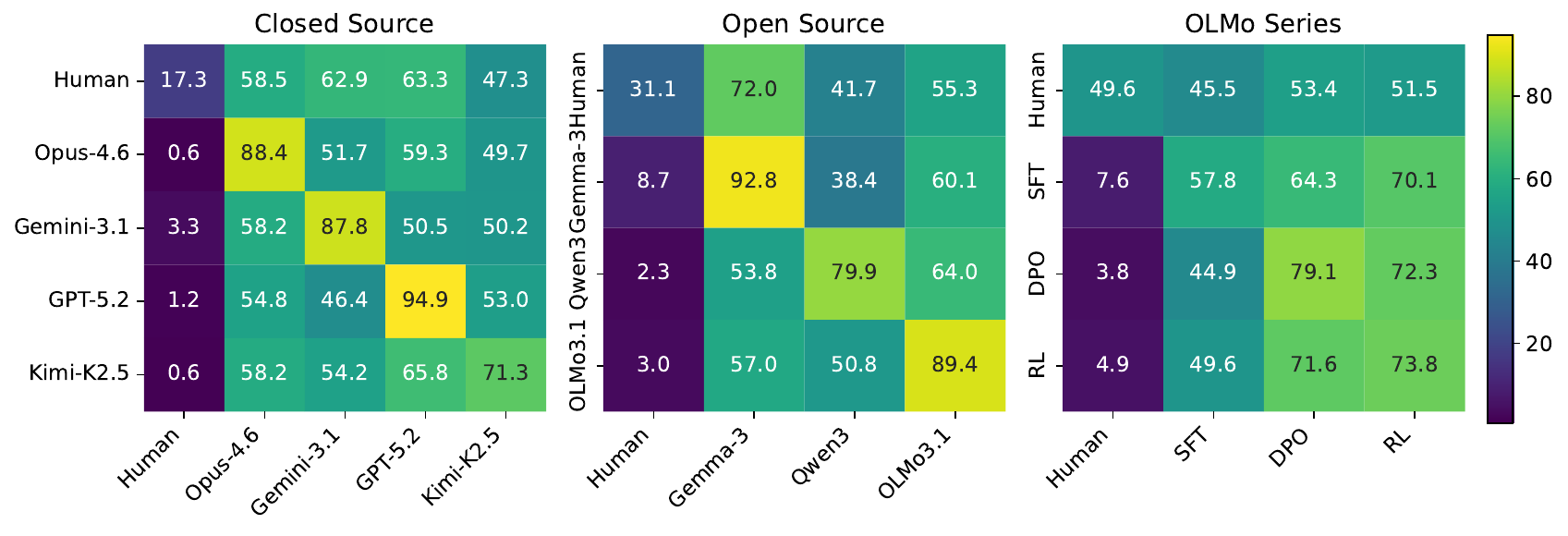}
    \caption{Similarity heatmap of triplets annotated by the LLM judge, showcasing the normalized selection rate within the closed-source, open-source, and OLMo generator pools without a confidence filter. The rows indicate the reference and the columns indicate the selected candidate generator, as detailed in \cref{sec:visualization}.}
    \label{fig:heatmap_main_noconf}
\end{figure*}

\begin{figure*}[ht]
    \centering
    \includegraphics[width=1.0\linewidth]{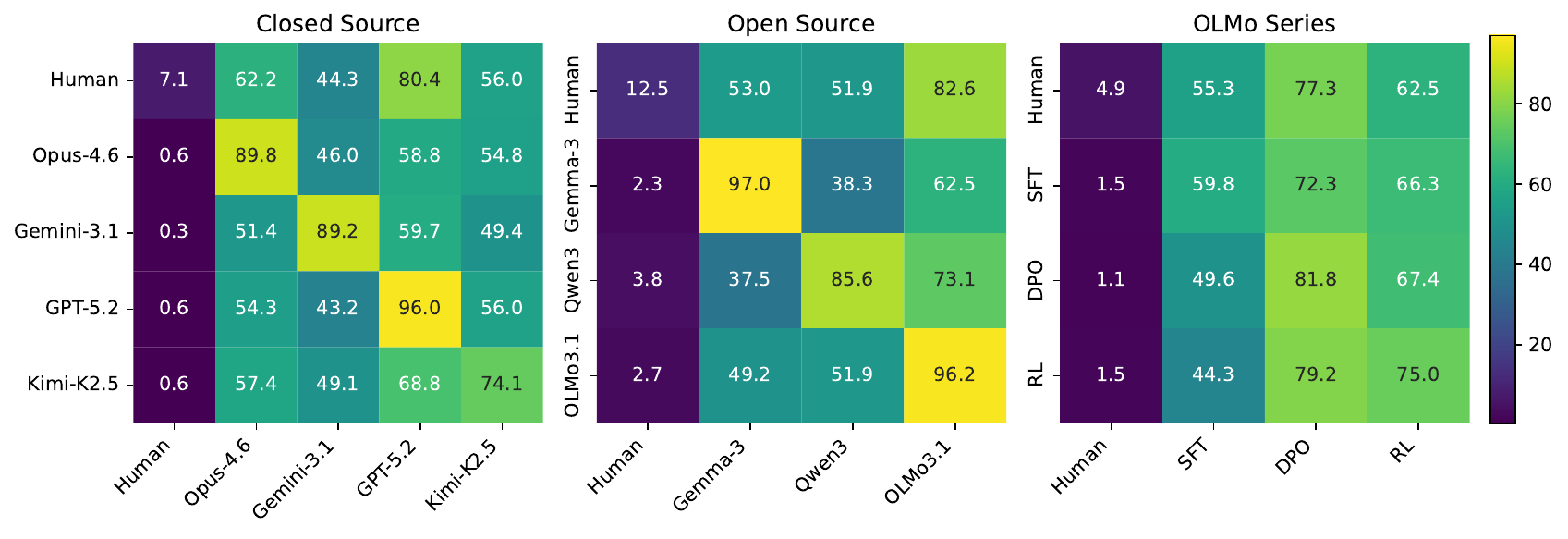}
    \caption{Similarity heatmap of triplets annotated by the preference model, within the closed source, open source, and OLMo pools.}
    \label{fig:heatmap_main_bt}
\end{figure*}

\begin{figure*}[ht]
    \centering
    \includegraphics[width=1.0\linewidth]{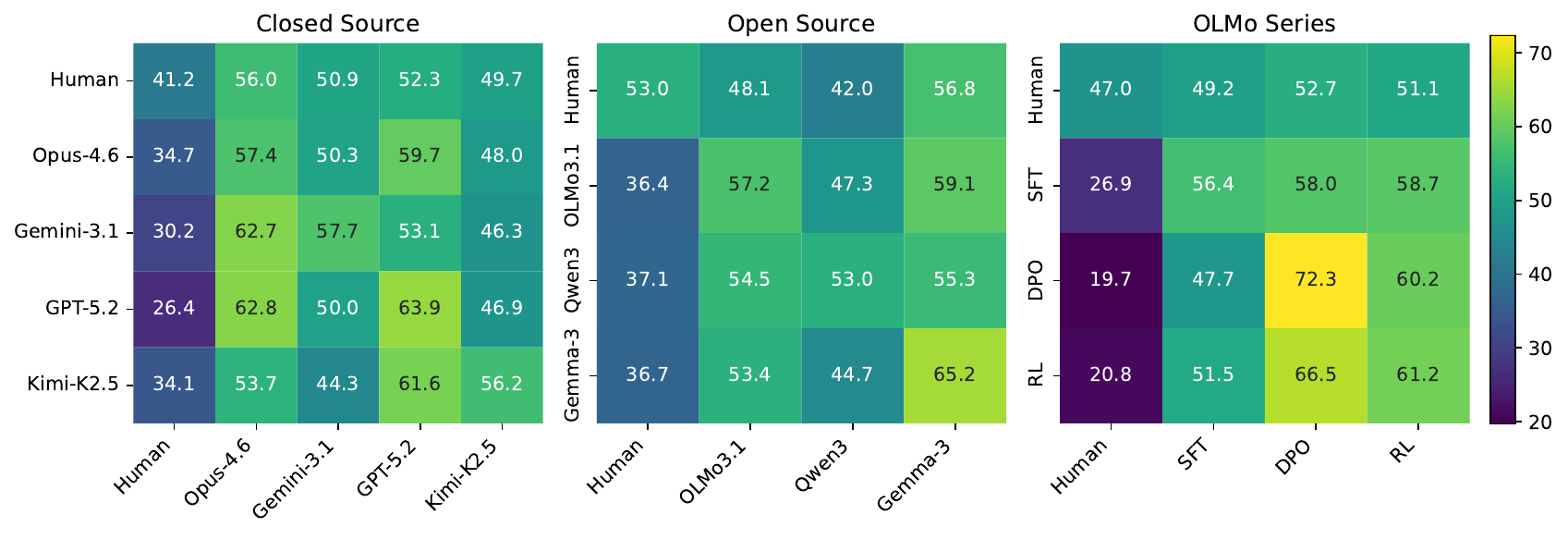}
    \caption{Similarity heatmap of triplets annotated by the LLM judge, showcasing the normalized selection rate within the closed-source, open-source, and OLMo generator pools under the \texttt{random} setting (all three stories in a triplet based on different writing prompts). The rows indicate the reference and the columns indicate the selected candidate generator, as detailed in \cref{sec:visualization}.}
    \label{fig:heatmap_main_random}
\end{figure*}

We provide additional details for all three automated annotation frameworks in this section.
\subsection{LLM-as-a-Judge}

The LLM-as-a-Judge setup directly utilizes an LLM to judge the similarity given a triplet. %Given the restricted nature of the task, we hypothesize that LLM-as-a-Judge will function better in this scenario than prior attempts that ask the judge to score diversity or creativity based on a rubric \citep{saakyanDeathNoveltyNGram2025, luchiniAutomatedAssessmentCreativity2025}. The results of the narrative similarity shared task also indicate that LLM-based systems are effective judging similarity in plot summaries, with up to $78\%$ accuracy \citep{hatzelSemEval2026Task4}.
Inspired by prior approaches to narrative evaluation, we decompose narrative similarity into several simple categories such as plot, themes, and characters \citep{paech2023eq,hatzelSemEval2026Task42026}. GPT-5 (\texttt{gpt-5}) is used as the judge model for all our experiments. The full prompt used for judging is given in \cref{prompt:prompt_llm_judge}.

\subsection{Narrative Component Embedding}
\label{sec:appendix:embedding}

For narrative component embedding, narrative components are extracted by an LLM from each story, and they are transformed into embeddings. Cosine similarity between the embeddings are then used to annotate the triplets. We use GPT-5 to extract the components, utilizing the framework opted by the top-scoring system in \citet{hatzelSemEval2026Task42026}, where each story is decomposed into overall abstract theme, course of action, and outcomes. We then use the Gemini Embedding (\texttt{gemini-embedding-001}) to embed all three components together as a concatenated string. The full prompt used for extracting narrative components is given in \cref{prompt:prompt_narrative_component_extraction}.

\subsection{Preference Model}
\label{sec:appendix:pm}

\begin{table}[h]
\centering
\begin{tabular}{lr}
\toprule
\textbf{Hyperparameter} & \textbf{Value} \\
\midrule
Learning Rate & $1 \times 10^{-5}$ \\
Batch Size & 2 \\
Training Epochs & 1.0 \\
Max Token Length & 8192 \\
\bottomrule
\end{tabular}
\caption{Hyperparameters used for training the preference model.}
\label{tab:pref_model_hyperparameters}
\end{table}

We utilize a discriminative Bradley-Terry (BT) formulation \citep{bradleyRankAnalysisIncomplete1952}, where in each annotated triplet $(s_r,s_s,s_d)$, the preference model $r_\theta$ learns to assign a higher scalar value to the more similar candidate $s_s$ than to the less similar candidate $s_d$ when evaluated against the reference $s_r$. The corresponding loss is defined as:

$$ \mathcal{L} = - \log \left( \sigma \left( r_\theta(s_r, s_s) - r_\theta(s_r, s_d) \right) \right) $$

We use Qwen3 1.7B (\texttt{Qwen/Qwen3-1.7B}) as our base model, appending a linear layer at the end and conducting full fine-tuning with the aforementioned formulation. We aggregate the GPT-5 annotations across the closed-source, open-source, and OLMo pools to create the training dataset. As we are using the human-annotated triplets for evaluation, we exclude all triplets with any overlap in stories. We reserve $2\%$ of the preference pairs to serve as an evaluation set and utilizing the remainder for training. We use AdamW as the optimizer. The hyperparameters used for training are provided in \cref{tab:pref_model_hyperparameters}. The model was trained on a single NVIDIA RTX A6000 48GB GPU, taking approximately 40 minutes to train. 

\section{Additional Results}
In this section, we provide additional results of the experiments described in the main text.

\subsection{Within Model Categories, Without Confidence Filter}
\label{sec:appendix:model_categories}

\cref{fig:heatmap_main_noconf} provide the resulting heatmaps from the comparison done in \cref{sec:results_main:model_categories}, but without the confidence filter applied. We note that all trends we point out in \cref{sec:results_main:model_categories} persist.

\subsection{Preference Model Heatmaps}
\label{sec:appendix:model_categories_bt}

\cref{fig:heatmap_main_bt} provides the resulting heatmaps of the setting described in \cref{sec:results_main:model_categories}, but annotated by the preference model instead of the LLM judge. We note that the overall trends remain, and the results are largely consistent with the LLM judge annotations, providing further evidence for its reliability.

\subsection{Sampling Parameter Sweep Heatmaps}
\label{sec:appendix:param_sweep}

\begin{figure*}[ht]
    \centering
    \includegraphics[width=1.0\linewidth]{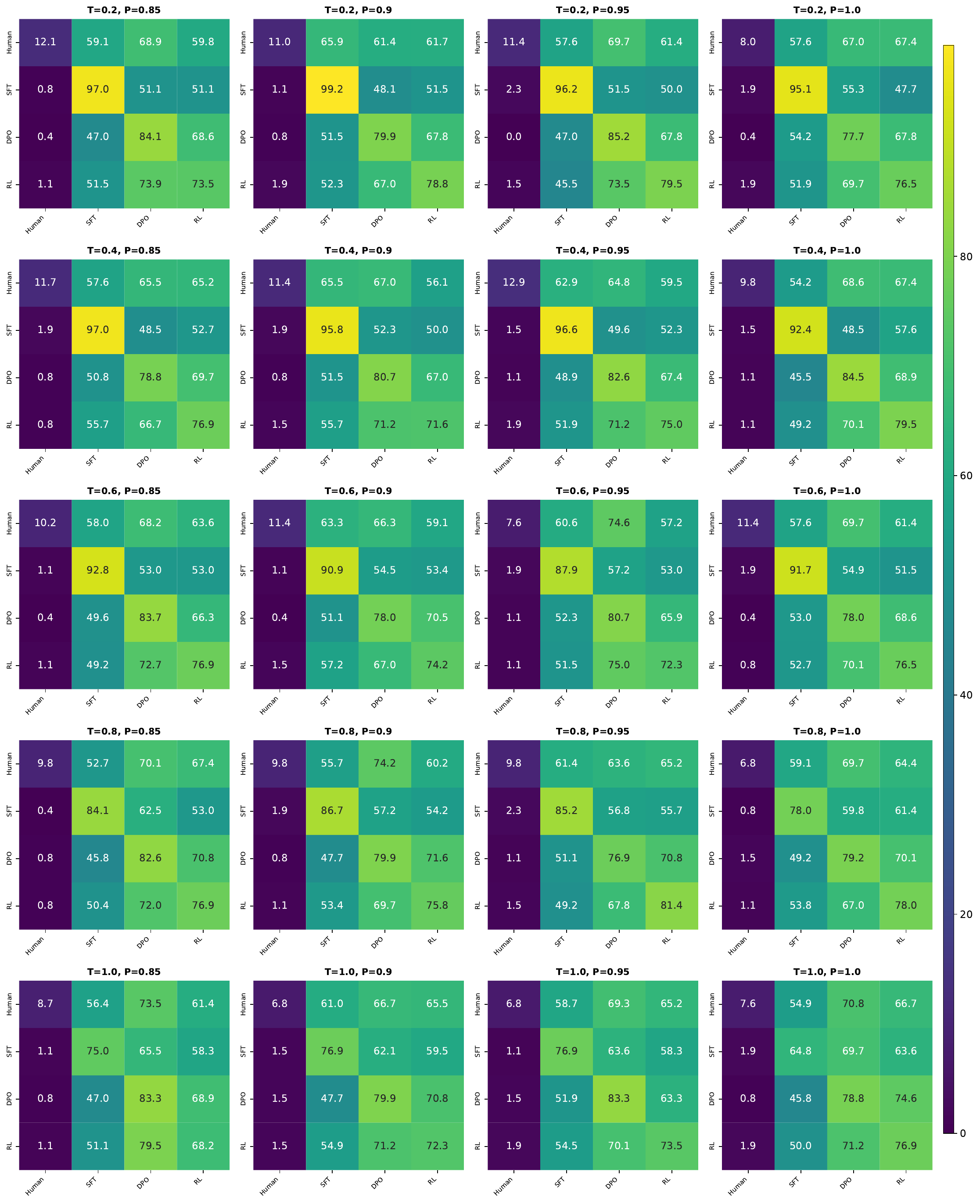}
    \caption{Similarity heatmap of triplets annotated by the preference model, within the OLMo pool under a sweep of sampling parameters, with all combinations of temperature $T\in \{0.2, 0.4, 0.6, 0.8, 1.0\}$ and top-P $P\in \{0.85, 0.9, 0.95, 1.0\}$.}
    \label{fig:heatmap_param_sweep}
\end{figure*}

\cref{fig:heatmap_param_sweep} provides the full heatmaps of the sampling parameter sweep described in \cref{sec:results_mitigation:sampling_params}. As we detailed, we find that top-P has no discernible effect on the resulting diversity, while temperature only has a notable effect on the OLMo SFT model.

\section{Additional Experiments}

In this section, we describe additional experiments that were omitted from the main text for brevity.

\subsection{Preference Model Evaluation with Narrative Component Extraction}
\label{sec:appendix:pm_component_evaluation}

As we directly train on LLM-annotated triplets, it is plausible that the preference model's scores may be based on other factors of similarity, such as lexical and syntactic distribution, which can coincide with narrative similarity in our dataset. To test this, we also conduct a separate evaluation where we remove all stylistic properties from the stories by extracting narrative components via GPT-5. We use the same prompt as narrative component embedding, extracting the story into overall abstract theme, course of action, and outcomes.
\begin{table}[ht]
\centering
\begin{tabular}{lcc}
\toprule
\textbf{Category} & \textbf{PM} & \textbf{PM Comp.} \\
\midrule
\texttt{(H, H, M\textsubscript{1})} & $72.7\%$ & $81.8\%$ \\
\texttt{(H, M\textsubscript{1}, M\textsubscript{2})} & $50.0\%$ & $31.8\%$ \\
\texttt{(M\textsubscript{1}, H, M\textsubscript{1})} & $100.0\%$ & $100.0\%$ \\
\texttt{(M\textsubscript{1}, H, M\textsubscript{2})} & $91.1\%$ & $83.9\%$ \\
\texttt{(M\textsubscript{1}, M\textsubscript{1}, M\textsubscript{2})} & $83.3\%$ & $83.3\%$ \\
\texttt{(M\textsubscript{1}, M\textsubscript{2}, M\textsubscript{3})} & $64.0\%$ & $72.0\%$ \\
\midrule
Overall & $77.0\%$ & $76.0\%$ \\
\bottomrule
\end{tabular}
\caption{Average preference model agreement with human annotations across different triplet categories, using the original stories (PM) as well as the agreement when using extracted narrative components (PM Comp.).}
\label{tab:preference_model_agreement}
\end{table}

We compare the preference model annotations against the human annotations, and compile the average agreement in \cref{tab:preference_model_agreement}. Both the component-extracted and original story evaluations report similar results, despite the component-extracted format not being included in the training data. This indicates that the preference model is robust to surface-level linguistic variations, and can likely be used as a general scoring model for similarity assessment.

\subsection{Random Comparison}
\label{sec:appendix:random}

In addition to the prompt-wise comparisons in \cref{sec:results_main}, we also test the case where all three stories are from different writing prompts. Like in \cref{sec:triplet_selection}, we iterate through all stories $s_r \in S$, with corresponding prompt $p_r$, and combinations of generators $\{a,b\} \in {G \choose 2}$. However, we then randomly sample from all stories generated by $a$ and $b$ respectively, such that all three are from mutually distinct prompts. As all three stories are written for different prompts, we hypothesize that the task becomes more ambiguous, and additional noise is introduced in the form of prompt selection. For example, in a triplet where the reference story prompt is distinctly more similar to one of the candidate story prompts, the narrative structure will likely follow suit regardless of the generators. Thus, we consider this setting to be less useful in making inferences, but still provide it for comparison.

We apply this setting for the same pools and annotation schemes as \cref{sec:results_main:model_categories}, with the resulting heatmaps compiled in \cref{fig:heatmap_main_random}. While the overall trends we pointed out in \cref{sec:results_main:model_categories} still remain, the magnitude is less pronounced, with LLM stories being judged to be more similar to a human story up to a $31.9\%$ of the time. As we detailed above, this setting is inherently more ambiguous with additional noise, and so the reduction in magnitude is to be expected. Nevertheless, even with completely different prompts, human-written stories are distinctly less similar to the LLM-generated ones.

\subsection{Chain-of-Thought}

\begin{figure}[ht]
    \centering
    \includegraphics[width=1.0\linewidth]{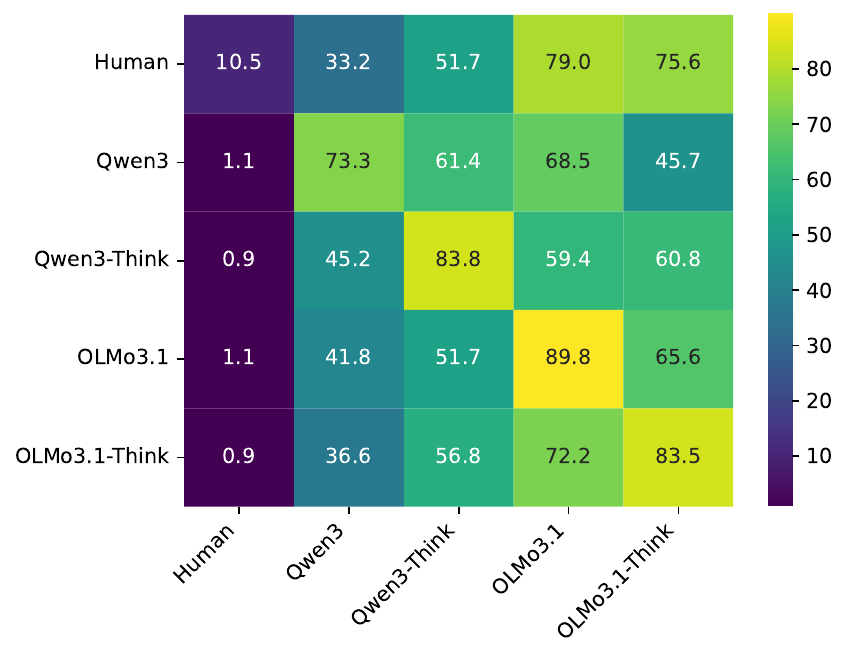}
    \caption{Similarity heatmap of triplets annotated by the preference model, showcasing the normalized selection rate comparing thinking and non-thinking variants of Qwen3 and OLMo3.1. The rows indicate the reference and the columns indicate the selected candidate generator, as detailed in \cref{sec:visualization}.}
    \label{fig:heatmap_cot_bt}
\end{figure}

We test whether chain-of-thought reasoning has any effect on narrative diversity by comparing the regular and chain-of-thought (or thinking) variants of Qwen3 and OLMo 3.1. For Qwen3 we simply enable and disable chain-of-thought via the system prompt, while for OLMo 3.1 we compare the OLMo 3.1 Thinking and OLMo 3.1 Instruct models. The results are compiled in \cref{fig:heatmap_cot_bt}. We note that with thinking enabled, Qwen3 has a higher self-similarity at $89.8$ vs $83.5$, but the OLMo Thinking variant has a lower self-similarity at $93.5$ vs $89.8$. In general, we find little meaningful differences overall, and so we conclude that chain-of-thought prompting may at best mildly affect the diversity, specific to the model family.

\section{AI Use}
\label{sec:appendix:ai_disclosure}

We utilize AI coding assistants (available via GitHub Copilot) in some cases to generate boilerplate code for experiments and data visualization. All generated code was reviewed and tested by the authors. We also use Gemini 3.1 Pro as a general-purpose assistant for formatting LaTeX elements and for checking grammar and typographical errors in the final drafts of the manuscript. No AI system was used to generate novel ideas or draw analytical conclusions, and no part of the manuscript was generated by such a system.

\begin{figure*}[ht]
\centering
\small
\begin{tcolorbox}[colback=white,colframe=black!75!white,colbacktitle=black!75!white,width=1.0\textwidth,title=Narrative Similarity Annotation Guidelines (1 of 2)]

{\normalsize \textbf{Task Overview}} \\
In this study, you are tasked with identifying similar stories. In each sample, three short stories of roughly 200-400 words will be provided: a reference story, story A, and story B. You will compare two candidate stories (story A and B) against the reference story and determine which candidate is more narratively similar to the reference. You will also be asked to rate your confidence in your judgement. All stories are written responses to the same writing prompt. Your annotations will be used for academic research.

\vspace{0.5em}
{\normalsize \textbf{Task Workflow}}
\begin{enumerate}
    \item Read the \textbf{Reference Story} carefully, noting its key narrative elements.
    \item Read \textbf{Story A} and \textbf{Story B} completely.
    \item Compare each candidate's narrative structure to the reference.
    \item Select which story (A or B) is more narratively similar.
    \item Rate your confidence (1-5) in this choice.
\end{enumerate}

\vspace{0.5em}
{\normalsize \textbf{Narrative Similarity}} \\
The narrative similarity of stories can be broken down into three core aspects: (1) the abstract themes of the story, (2) the course of action, and (3) the story outcomes. At one extreme, this means that the story deals with the same themes and tells the same order of events with an identical outcome or conclusion, just using a different wording; at the other extreme, the story might be completely different and lack any basis for comparison. \\\\
More difficult to assess are stories that only share some similarities. In such cases, you are asked to weigh the three core components of story similarity. You should focus on the core aspects of stories, potentially largely ignoring side storylines. How you weigh the individual factors should be based on your intuitive impression of which aspects you consider crucial to the overall similarity of the specific stories. \\\\
We define these three aspects as follows: 
\begin{itemize}
    \item \textbf{Abstract Theme} describes the defining constellation of problems, central ideas, and core motifs of a story. The definition does not cover the concrete setting of a story.
    \item \textbf{Course of Action} describes sequences of events, actions, conflicts, and turning points in a story and the order in which they happen.
    \item \textbf{Outcomes} describe the results of the plot at the end of the text, for example, the conflict resolution, the characters' fates, moral lessons, etc. It does not cover intermediate statuses that change later in the story.
\end{itemize}

Each aspect can take different forms in an actual pair of stories. Below, we list one example for each aspect:
\begin{itemize}
    \item The general setting of the story, if it strongly influences the events in the story or the events necessitate a specific setting (abstract themes)
    \begin{itemize}
        \item A: On the week-long journey from Europe to the Americas, the crew members get into a heated conflict about the best ration packages.
        \item B: The flight to Mars is long. After several weeks, the astronauts become better friends than ever before, having to share the limited resources.
        \item A and B share some similarities in that the polar opposite outcomes are both enabled by being cut off from the outside world.
    \end{itemize}
    \item The order of events in the story (course of action)
    \begin{itemize}
        \item A: After the ship capsizes and Alice barely makes it out alive, she starts living life to the fullest.
        \item B: Alice is living life to the fullest until, one day, her ship capsizes. She barely makes it out alive.
        \item A and B are similar in that both tell of a good life and a shipwreck (abstract theme), but they differ in the course of action, and the order is very different.
    \end{itemize}
    \item The outcomes of events (story outcomes)
    \begin{itemize}
        \item A: The man intentionally drops a cup; it breaks.
        \item B: He accidentally swipes the bottle off the table, and it shatters.
        \item A and B are similar in that the events are comparable and lead to similar outcomes.
    \end{itemize}
\end{itemize}

\end{tcolorbox}
\caption{The annotation guidelines provided to human annotators (part 1 of 2).}
\label{prompt:narrative_similarity_guidelines_1}
\end{figure*}

\begin{figure*}
\centering
\small
\begin{tcolorbox}[colback=white,colframe=black!75!white,colbacktitle=black!75!white,width=1.0\textwidth,title=Narrative Similarity Annotation Guidelines (2 of 2)]
        
There are a range of factors that expressly do NOT contribute to the narrative similarity:
\begin{itemize}
    \item The style of writing in a story
    \item The concrete setting of a story (also including the time period).
    \item The names of the characters and locations
    \item The length of a text
    \item The level of detail in which the events are told
\end{itemize}

\vspace{0.5em}
{\normalsize \textbf{Differentiating Between Similarity Aspects}} \\
Distinguishing the three aspects can be challenging. In general, it is important to consider each aspect independently. \\\\
Often, pairs of stories that are similar in terms of course of action will also share an abstract theme. However, it is possible that similar events emerge from completely different surrounding circumstances. Outcomes, on the other hand, are clearly distinct from the other two aspects: practically identical events in stories with comparable abstract themes can result in polar opposite outcomes. \\\\
When comparing abstract themes, it can help to explicitly formulate them. There is, of course, no single correct answer, and a single story's theme could be formulated in many ways. Two stories share a general theme if there is a description that captures the defining circumstances of both stories.

\vspace{0.5em}
{\normalsize \textbf{Confidence Scale}} \\
When selecting the confidence, it is important to note that it should not necessarily be the amount of similarity between the reference and chosen candidate story, but rather your confidence that the chosen candidate story is more similar to the reference than the other candidate story. In specific, each score can be interpreted as written below: 
\begin{itemize}
    \item \textbf{1}: No confidence; Both stories A and B are equally similar to the reference story in narrative
    \item \textbf{2}: Slight confidence; The similarities are comparable, and it could go either way
    \item \textbf{3}: Moderate confidence; One story is noticeably more similar to the reference, but there is room for doubt
    \item \textbf{4-5}: Strong confidence; One story is clearly and distinctly more similar to the reference
\end{itemize}

\vspace{0.5em}
{\normalsize \textbf{Example}} \\
In the following example, summaries are used for faster reading, but the task itself will use the full short stories. \\\\
\textbf{Reference}: Mara writes a poignant, unsent letter to her deceased brother, Noah, reminiscing about their shared childhood and updating him on her life. She ultimately places the letter in a drawer filled with dozens of others, illustrating her quiet, ongoing process of grieving and holding onto his memory. \\
\textbf{Story A}: Margaret writes an anguished letter to Steve, longing for the ``faithful, loyal man'' she once had and expressing deep guilt over trying to replace him with someone who ultimately broke her trust. It is only in the final sign-off that the true nature of their relationship is revealed; the writer is a daughter mourning the father she lost nineteen years ago. \\
\textbf{Story B}: A young man drafts a nostalgic letter to her friend, Sarah, recalling their vibrant past and the small, beautiful details of their shared life. Even though Sarah is in a facility suffering from severe dementia and may not understand the words, he mails the letter anyway, deciding that love does not require memory to exist. \\\\
\textbf{Choice}: Story B, Confidence 5 \\\\
All three stories are nostalgic letters reflecting on the shared history of a lost loved one. Story A diverges from the reference by using misdirection to imply that the letter is initially to a lover, and includes an exploration of her guilt over ``trying to replace him''. While Story B diverges in some surface details (young man writing instead of a named woman, writing to a friend instead of family), it shares the reference's straightforward focus on a deeply rooted childhood bond. Thus, Story B is more narratively similar.

\end{tcolorbox}
\caption{The annotation guidelines provided to human annotators (part 2 of 2).}
\label{prompt:narrative_similarity_guidelines_2}
\end{figure*}

\begin{figure*}[ht]
\centering
\small
\begin{tcolorbox}[colback=white,colframe=black!75!white,colbacktitle=black!75!white,width=1.0\textwidth,title=Prompt for LLM-as-a-Judge triplet annotation.]
You are an expert literary analyst. Your task is to critically evaluate narrative similarity between two candidate stories and a reference story.

You will be given:
\\- A reference story
\\- Two candidate stories (Writer A \& Writer B)

Your job is to determine which candidate story is more similar to the reference in several narrative criteria.

[REFERENCE STORY]

\{story\_ref\}

[/REFERENCE STORY]
[WRITER A]

\{story\_a\}

[/WRITER A]
[WRITER B]

\{story\_b\}

[/WRITER B]

Judge which writer's story more closely matches the reference story on each of these dimensions:

- Plot similarity (events, structure, narrative arc)
\\ - Character similarity (roles, motivations, arcs, emotional dynamics)
\\ - Setting similarity (time, place, atmosphere)
\\ - Tone \& mood similarity (emotional feel, narrative voice)
\\ - Theme similarity (underlying message, core concepts)
\\ - Style similarity (pacing, structure, point of view)
\\ - Overall similarity (general closeness to reference story)

Judging notes:

- Surface details (e.g., character names, specific locations, diction) do not matter; focus on the narrative elements.
\\- Be aware that these similarities may be independent, i.e. a story may be more similar to the reference in one and less in another.
\\- Outputs will sometimes be truncated to ensure length consistency. Don't penalise this, just judge what is there on its merit.
\\- You must always pick a winner for each criteria (no draws)

For each criterion, assign a winner (Writer A or Writer B), like so:
\begin{verbatim}
"plot_similarity": "A" 
"character_similarity": "B"
"setting_similarity": "A"
\end{verbatim}

In addition to picking a winner, you should also assign a confidence rating for the overall similarity, which should reflect the confidence in disambiguation. The score should be between 1 (both stories extremely similar, too similar to consistently tell apart) to 5 (the winning story is explicitly and concretely more similar).

Respond in valid json without additional commentary (remembering to escape any string quotes), in this format:
\begin{verbatim}
{
"analysis": "detailed analysis and reasoning about the coming scoring decisions",
"plot_similarity": "winner & disparity rating",
"character_similarity": "winner & disparity rating",
"setting_similarity": "winner & disparity rating",
"tone_mood_similarity": "winner & disparity rating",
"theme_similarity": "winner & disparity rating",
"style_similarity": "winner & disparity rating",
"overall_similarity": "winner & disparity rating",
"confidence": "confidence rating for overall similarity"
}
\end{verbatim}
\end{tcolorbox}
\caption{Prompt for LLM-as-a-Judge triplet annotation.}
\label{prompt:prompt_llm_judge}
\end{figure*}

\begin{figure*}[ht]
\centering
\begin{tcolorbox}[colback=white,colframe=black!75!white,colbacktitle=black!75!white,width=1.0\textwidth,title=Prompt for Narrative Component Extraction]

Describe a given story in JSON format. You need to describe the following three components: 
\\\\
1) Overall abstract theme: Describe in brief the central ideas, core motifs and defining constellation of problems. For example, in both these stories: 

A: "On the week-long journey from Europe to the Americas, the crew members get into a heated conflict about the best ration packages." 

B: "The flight to Mars is long. After several weeks, the astronauts become better friends than ever before, having to share the limited resources." 
Theme: A story about people isolated from outside world in a journey, and how it affects their interpersonal relationship. 
\\\\
2) Course of action/events: Describe in brief the sequence of events that actually happens in the story. For example, in the following stories: 

A: "After the ship capsizes and Alice barely makes it out alive, she starts living life to the fullest with a new-found perspective about how precious life is." Events: Alice’s ship capsizes. Alice barely makes it out alive. Alice starts living life to the fullest. 

B: "Alex loses his engagement ring while swimming. He freaks out, and after hours of diving for it, he still cannot find it." 
Events: Alex loses his engagement ring while swimming. Alex freaks out. Alex looks for it. Alex fails to find it. 
\\\\
3) The outcomes: Describe in brief the final ending or outcomes of the story. For example, in the following stories: 

A: "Anna loses her purse. She retraces her steps but cannot find it. Dan finds it and helpfully returns it to her." 
Outcome: Someone finds a lost item and returns to owner. 

B: "Brian lost his backpack. He was terrified because there were important documents in it. After an hour of intense search he finally found it." 
Outcome: Someone finds their lost item. 

C: "Jill was driving home when another car suddenly crashed into hers. After receiving medical attention, she recovered within just days and now advocates for traffic safety." 

Outcome: A person advocates for traffic safety after recovering from car crash. 
\\\\
You should produce a valid JSON object with the three attributes describing the given story: "theme", "events" and "outcome". Do not produce any extra explanation or additional text.

\end{tcolorbox}
\caption{Prompt for Narrative Component Extraction.}
\label{prompt:prompt_narrative_component_extraction}
\end{figure*}

\begin{figure*}[ht]
\centering
\begin{tcolorbox}[colback=white,colframe=black!75!white,colbacktitle=black!75!white,width=1.0\textwidth,title=Prompts for Sequential Generation]

\textbf{System Prompt}

You are a creative writer tasked with writing stories based on given prompts and 
word count requirements. Do not reply with anything other than the story itself. 
Do not include any commentary, explanations,  or notes, just the story.

\vspace{0.5em}

\textbf{User Prompt}

Write a short story between \{min\_words\} and \{max\_words\} words based on the writing prompt: \{wp\}

\vspace{0.5em}

\textbf{Follow-up User Prompt for Sequential Negative Generation}

Generate another story based on the same writing prompt and word range. 
Try to make it as different from previous stories as possible.

\end{tcolorbox}
\caption{All LLM prompts used for generating stories.}
\label{prompt:sequential_generation}
\end{figure*}

\end{document}